\documentclass{opendatalab}

\usepackage[utf8]{inputenc}
\usepackage{xcolor}
\usepackage[most]{tcolorbox}
\usepackage{longtable}
\usepackage{listings}
\usepackage[numbers]{natbib}
\usepackage{enumitem}
\usepackage{graphicx}
\usepackage{xspace} 
\usepackage{hyperref}
\usepackage{xspace}
\usepackage[table]{xcolor}
\definecolor{baselinegray}{HTML}{EEF2F7}
\newcommand{\basecell}[1]{\cellcolor{baselinegray}{#1}}
\makeatletter
\DeclareRobustCommand\onedot{\futurelet\@let@token\@onedot}
\def\@onedot{\ifx\@let@token.\else.\null\fi\xspace}

\makeatother
\definecolor{codegreen}{rgb}{0,0.6,0}
\definecolor{codegray}{rgb}{0.5,0.5,0.5}
\definecolor{codepurple}{rgb}{0.58,0,0.82}
\definecolor{backcolour}{rgb}{0.95,0.95,0.92}
\definecolor{promptcolor}{HTML}{D1D0F2}
\definecolor{promptcolorheader}{HTML}{bdbcec}

\definecolor{promptcolor}{HTML}{D1D0F2}
\definecolor{promptcolorheader}{HTML}{bdbcec}

\newtcolorbox{promptbox}[1][]{
  enhanced, breakable,
  top=0.3em,bottom=0.3em,left=0.5em,right=0.5em,
  toptitle=0.3em,bottomtitle=0.2em,boxsep=0pt,
  colframe=promptcolorheader, colback=promptcolor!50, boxrule=0.5pt,
  width=\columnwidth, 
  title={\footnotesize #1} 
}
\lstdefinestyle{promptstyle}{
    backgroundcolor=\color{backcolour},   
    commentstyle=\color{codegreen},
    keywordstyle=\color{magenta},
    numberstyle=\tiny\color{codegray},
    stringstyle=\color{codepurple},
    basicstyle=\ttfamily\footnotesize,
    breakatwhitespace=false,         
    breaklines=true,                 
    captionpos=b,                    
    keepspaces=true,                 
    numbers=left,                    
    numbersep=5pt,                  
    showspaces=false,                
    showstringspaces=false,
    showtabs=false,                  
    tabsize=2
}
\title{SPARK: Susceptibility-Guided Profiling and Steering of Latent Reasoning States in Large Language Models}

\author{
    \parbox{\linewidth}{\raggedright
        Dongxu Zhang$^{1}$ \quad
        Yiding Sun$^{1}$ \quad
        Zihao Guo$^{1}$ \quad
        Xiangyang Yang$^{1}$ \quad
        Kai Tang$^{2}$ \\ \vspace{0.15cm} 
        Lin Chen$^{1}$ \quad
        Cheng Tan$^{3}$ \quad
        Jihua Zhu$^{1}$ \\ \vspace{0.15cm}
    }
}

\affiliation{
    \vspace{0.2cm} 
    \parbox{\linewidth}{\raggedright \small
        $^1$Xi'an Jiaotong University \quad
        $^2$Peking University \quad
        $^3$Tencent \\ \vspace{0.1cm}
    }
}

\abstract{
Reasoning failures in large language models (LLMs) are usually evaluated from final answers, but a wrong answer does not reveal why the model failed. The same incorrect output may reflect missing capability, an unstable reasoning trajectory, or a failure to activate a reasoning state that is already available in the frozen model. Existing prompting and benchmark-based evaluation methods mostly operate at the output level, while generic activation-steering methods typically apply global directions without diagnosing which examples require intervention. In this paper, we introduce \textsc{SPARK}, which uses hidden-state response to diagnose whether a model internally enters an effective reasoning state and to guide lightweight test-time steering. The key observation is that raw hidden-state susceptibility is strongly confounded by prompt length, especially in programmatic and algorithmic reasoning where harder serialized instances naturally become longer. \textsc{SPARK} therefore uses length-controlled susceptibility to separate input-scale effects from residual reasoning activation, and combines this signal with cross-layer coordination to select reasoning-active anchors and under-activated hard examples. We use \textsc{FRONTIER}-4.5K as a controlled programmatic reasoning suite for latent profiling and difficulty-aware analysis, and evaluate \textsc{SPARK}-Steering on GSM8K and MATH-500 with forward-only benchmark profiling. Our method improves Qwen3 series models consistently; on MATH-500, accuracy rises from 82.0\% to 84.6\% for Qwen3-4B and from 82.4\% to 85.6\% for Qwen3-8B. These results suggest that susceptibility can serve not only as a diagnostic signal for reasoning failures, but also as a practical guide for targeted test-time intervention.

\noindent \textbf{Keywords:} Large Language Models, Chain-of-Thought, Latent Reasoning States}

\metadata[Contact]{Dongxu Zhang, \email{zhangdongxu@stu.xjtu.edu.cn} 
\\
\\
}

\begin{document}

\maketitle


\begin{figure*}[t]
  \centering
  \includegraphics[width=0.85\textwidth]{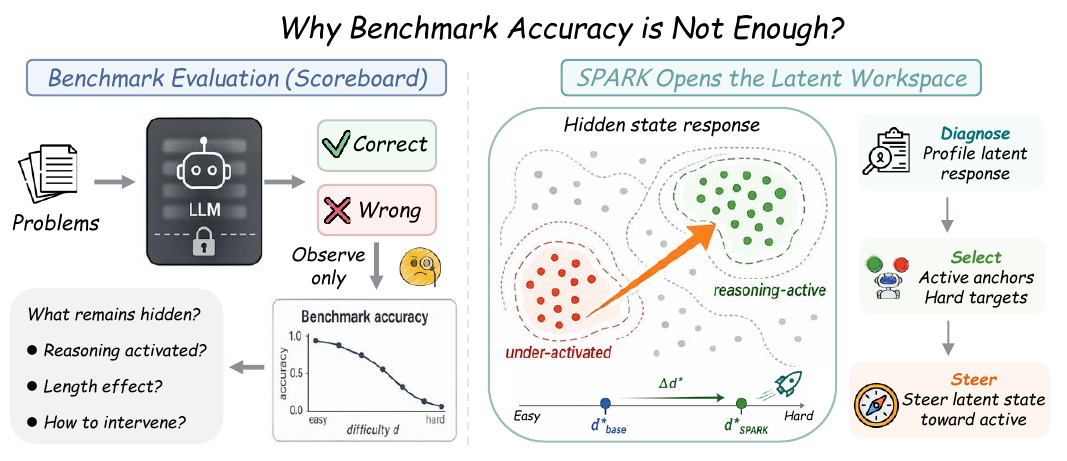}
  \caption{\textbf{Motivation and overview of \textsc{SPARK}.} Benchmark evaluation observes correct and wrong answers and summarizes performance with accuracy curves, but leaves the latent reasoning state hidden. \textsc{SPARK} complements this output-level view with latent profiling and test-time steering. It diagnoses hidden-state response, selects reasoning-active anchors and under-activated hard targets, and evaluates whether steering improves harder examples while preserving easier cases.}
  \label{fig:intro}
\end{figure*}

\section{Introduction}
\label{sec:int}
Large language models (LLMs) have made rapid progress on mathematical, symbolic, and algorithmic reasoning tasks~\cite{wu2026intern,xu2026nanoresearch,huang2026cfms}, yet their reasoning failures remain difficult to interpret from final answers alone~\cite{zhang2025ascot}. A model may solve easy instances reliably, become unstable on moderately difficult inputs, and fail sharply as problem structure grows~\cite{tang2026mitigatinghallucinationsinterlayerconsistency,tang2026seememitigatinghallucinationslarge,guo2026fademitigatinghallucinationsreducing}. Benchmark accuracy captures this behavioral pattern only at the output level~\cite{srivastava2023beyond,liang2022holistic}. It tells us when performance drops, but not what internal condition has changed. This matters because two incorrect answers may correspond to different internal states. In one case, the model may not possess the required capability~\cite{hendrycks2021measuring}. In another, it may have partial capability but fail to activate the latent computation needed for the current input~\cite{wei2022emergent}. Moreover, hard reasoning problems often become longer and more structurally dense~\cite{zhang2026chain}, so weak hidden-state response may reflect input scale rather than a true collapse of reasoning.

Current evaluation practice creates a gap between observing failure and deciding how to intervene. Accuracy curves can locate a behavioral capability boundary, but they do not specify what should be changed when a hard example fails. Prompting treats the model largely as a black box~\cite{wei2022chain,kojima2022large}, while fine-tuning changes parameters and obscures whether the frozen model already contains a usable reasoning state~\cite{hu2022lora,wang2026pointrft,sun2026tri}. A useful reasoning diagnostic should therefore be measurable during a forward pass, require no parameter updates, and provide an actionable signal for targeted test-time intervention.

We address this problem through hidden-state susceptibility. Inspired by susceptibility as response to a small perturbation~\cite{goodfellow2014explaining,novak2018sensitivity}, we measure how strongly transformer hidden states react when small noise is applied to the input embeddings. This yields a lightweight response signal for each problem, with the initial intuition that examples near a behavioral boundary may show amplified or unstable internal responses. Our experiments show that this intuition must be refined. In programmatic reasoning, raw susceptibility is strongly affected by prompt length~\cite{press2021train,liu2024lost}, especially in algorithmic tasks where harder graph-structured instances naturally become longer after serialization. As difficulty increases, raw hidden-state response can decrease largely because the input is longer; without length control, a metric that appears to describe reasoning may instead describe input scale.

This observation motivates a length-controlled view of susceptibility. Instead of asking whether a problem has high or low raw response, we compare its response with what is expected among inputs of similar length. The resulting residual signal better reflects reasoning activation after removing a major input scale confound. A positive residual suggests stronger internal activation than length alone would predict, while a negative residual suggests under activation. This distinction is especially important for hard failures, where weak raw response could otherwise be mistaken for a trivial consequence of long prompts.
We present \textsc{SPARK} for susceptibility guided profiling and steering of latent reasoning states. Following the pipeline in Figure~\ref{fig:intro}, the framework first estimates the behavioral boundary from the accuracy curve over difficulty. It then profiles latent states~\cite{geva2021transformer,meng2022locating} using length controlled susceptibility and cross-layer coordination, separating reasoning-active examples from hard examples that remain under activated after length correction. This partition is based on corrected internal evidence rather than prompt length or raw response, and turns latent profiling into a selection mechanism for intervention~\cite{belrose2023eliciting,ghandeharioun2024patchscopes}.

This diagnosis further enables \textsc{SPARK}-Steering, a targeted test-time intervention for frozen models~\cite{rimsky2024steering}. Instead of applying a generic activation direction, \textsc{SPARK} extracts a reasoning direction from active anchors, representing the latent shift from a baseline state toward an active reasoning state. During inference, the direction is injected into selected layers~\cite{turner2023steering,zou2023representation} for under-activated hard targets. This makes steering diagnosis-guided and selective, it aims to repair hard failures while preserving easier cases, without updating model parameters.

We evaluate \textsc{SPARK} on \textsc{FRONTIER}-4.5K, a 4,500-instance programmatic reasoning suite covering symbolic composition, logical inference, and algorithmic reasoning. These tasks allow us to compare behavioral accuracy with internal response as difficulty changes. Algorithmic reasoning is the primary diagnostic setting because it exhibits both a clear capability boundary and a strong length confound. In this paper, our contributions are as follows:
\begin{itemize}
\item We construct \textsc{FRONTIER}-4.5K, a reasoning benchmark with executable supervision and continuous difficulty annotations for joint behavioral and latent-state analysis.

\item We identify length as a critical confound in hidden-state susceptibility analysis, showing that raw response signals can reflect input-scale attenuation rather than reasoning activation, especially in algorithmic reasoning.

\item To the best of our knowledge, we present the first susceptibility-guided steering framework for implicit reasoning states. \textsc{SPARK} extracts reasoning directions from active anchors and selectively injects them into under-activated hard examples at test time.

\item Extensive experiments show that \textsc{SPARK} improves GSM8K, MATH-500, and \textsc{FRONTIER}, with stronger gains on harder reasoning cases and the importance of susceptibility-guided direction and layer selection.

\end{itemize}

\section{Related Work}
\subsection{Reasoning in Large Language Models.}
LLMs have shown strong performance on mathematical, symbolic, and commonsense reasoning benchmarks, while their success often changes sharply with model scale, prompt design, and task complexity~\cite{zhang2026pointcot,hendrycks2021measuring,cobbe2021training}. A large body of work improves reasoning by eliciting intermediate computations in the output space. Chain-of-thought (CoT) prompting encourages models to generate step-by-step rationales \citep{wei2022chain}, zero-shot variants show that such behavior can be triggered by simple natural-language instructions~\cite{kojima2022large}, and self-consistency improves robustness by marginalizing over multiple sampled reasoning paths~\citep{zhang2026not}. Subsequent prompting methods further decompose hard problems into simpler subproblems or search over intermediate reasoning states, such as least-to-most prompting, Tree of Thoughts, and ReAct-style reasoning with actions~\cite{zhou2022least,yao2023tree,yao2022react}. Other work studies verifier-guided reasoning and process supervision, emphasizing that correct final answers may depend on the stability of intermediate reasoning trajectories~\cite{zelikman2022star,lightman2024let}. These methods primarily operate through prompts, demonstrations, decoding, or supervision over generated rationales. In contrast, SPARK does not primarily aim to elicit longer or more explicit reasoning traces; it diagnoses whether the model internally enters a reasoning-active state.

\subsection{Activation Steering and Test-Time Control.}
Another line of work controls language models at inference time without updating their parameters. Early approaches guide generation through external discriminators, gradients, or future-token scoring~\cite{yang2021fudge,krause2021gedi,dathathri2019plug}. More recent activation-level methods directly modify internal states using directions derived from contrastive prompts or representation analyses~\cite{li2023inference,rimsky2024steering}. Activation addition, representation engineering, inference-time intervention, and contrastive activation addition show that adding a vector to selected hidden layers can steer attributes such as sentiment, honesty, toxicity, or refusal behavior without full fine-tuning~\cite{turner2023steering,zou2023representation}. These results suggest that some model behaviors are organized along manipulable directions in activation space. SPARK-Steering applies this idea to reasoning, but differs in two ways. First, the steering direction is not chosen from a generic semantic contrast; it is extracted from reasoning-active anchors selected by length-controlled susceptibility and cross-layer coordination $\Phi$. Second, the target of intervention is not a global style or safety attribute, but a specific failure mode: high-difficulty examples whose hidden-states are under-activated relative to same-length inputs. This connects latent diagnosis with a causal test-time intervention for reasoning.

\section{Reasoning Benchmark}

Our analysis requires a benchmark that exposes more than final answers. To study behavioral boundaries together with latent activation, each instance should provide verified supervision, a controlled difficulty coordinate, and metadata about its generated structure. We construct \textsc{FRONTIER}-4.5K, a programmatic reasoning benchmark with 4,500 generated instances for evaluating both output accuracy and hidden-state response. \textsc{FRONTIER}-4.5K contains three domains with 1,500 instances each. Symbolic Composition requires multi-step transformations over symbols and variable bindings; Logical Inference requires multi-hop rule chaining with distractor facts; and Algorithmic Reasoning requires graph- and procedure-based queries such as connectivity, path existence, and shortest-path problems~\cite{velivckovic2022clrs,markeeva2024clrs}. All answers are produced by deterministic solvers, avoiding model-based judging.

Each instance includes a problem text $x$, verified answer $y$, domain label, generation parameters, token length $T$, and normalized difficulty $d\in[0,1]$. Difficulty is derived from domain-specific generation factors and normalized within each domain to ensure comparable ordering: composition depth and variable interactions for symbolic tasks, rule-chain length and distractors for logical tasks, and graph size, edge density, and query complexity for algorithmic tasks.

\section{Method}

\subsection{Problem Setup}

Let $\mathcal{D}=\{(x_i,y_i,d_i)\}_{i=1}^{N}$ be a domain-specific split from one downstream benchmark. Each input $x_i$ has a verified answer $y_i$ and a normalized difficulty score $d_i\in[0,1]$, where larger values indicate harder instances.  We use this set to estimate behavioral boundaries, fit length-control models, and select reasoning-active anchors. A frozen language model $f_\theta$ produces an answer $\hat y_i$, and correctness is denoted by:
\begin{equation}
    c_i = \mathbf{1}[\hat y_i = y_i].
\end{equation}
The token length of $x_i$ is denoted by $T_i$. Throughout this section, model parameters $\theta$ are fixed. All profiling signals are computed from forward passes only.

The method has two stages. First, \textsc{SPARK Profiling} estimates the behavioral boundary and diagnoses latent reasoning states using hidden-state response. Second, \textsc{SPARK-Steering} uses the diagnosed active states to construct a test-time steering direction for hard under activated examples.

\subsection{Behavioral Capability Boundary}

The behavioral boundary is estimated from the relation between difficulty and correctness. We fit an accuracy curve:
\begin{equation}
    \hat p(d) \approx P(c=1\mid d),
\end{equation}
using held out calibration examples. The capability boundary $d^\star$ is defined as the difficulty level where the fitted pass rate drops to 50 percent, $d^\star =\inf\{d\in[0,1]\mid \hat p(d)\le 0.5\}.$ This boundary is a behavioral quantity. It is estimated from correctness, not from hidden-state metrics.

\begin{figure*}[t]
  \centering
  \includegraphics[width=1.0\textwidth]{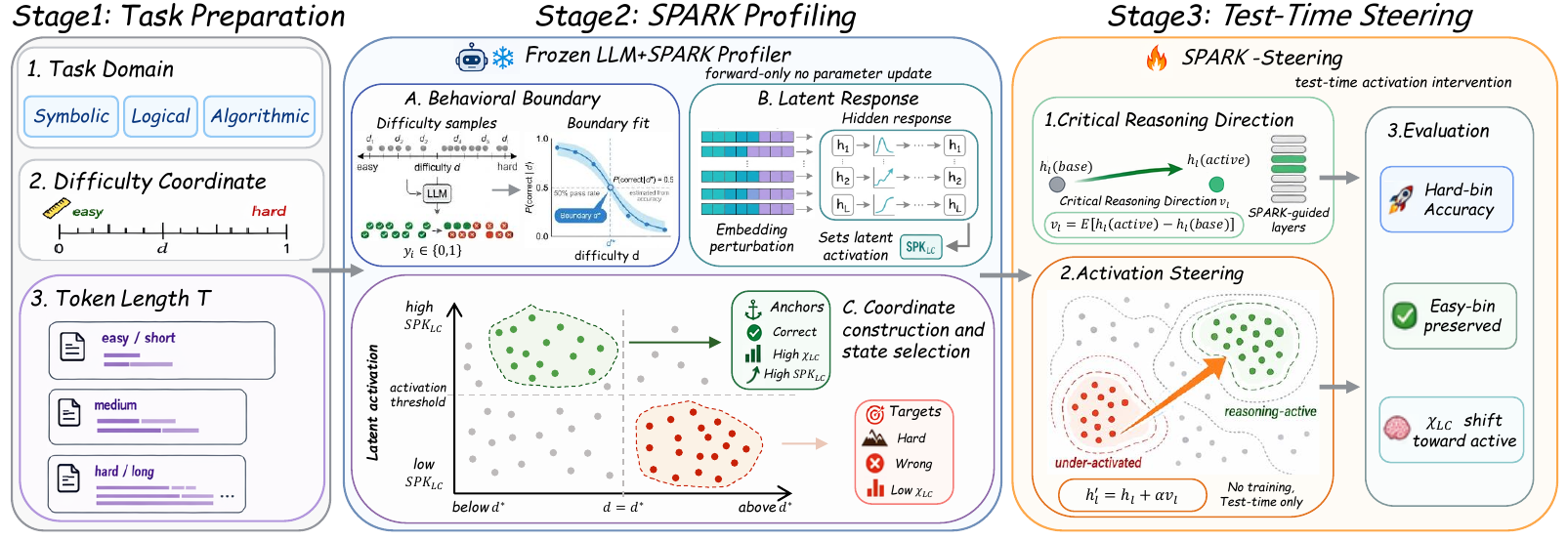}
  \caption{\textbf{Overview of \textsc{SPARK}.} Given tasks organized by domain, difficulty, and token length, SPARK first profiles a frozen LLM to identify the behavioral boundary and latent activation patterns associated with successful reasoning. It then selects active anchors and under-activated hard examples to estimate critical reasoning directions, which are injected at test-time through activation steering. The resulting intervention is applied only at inference time.}
  \label{fig:method}
\end{figure*}

\subsection{Latent Susceptibility}

For an input $x$, let $E(x)$ be its input embedding sequence. For layer $l$, let $r_l(x)$ be the representation used for latent analysis, $r_l(x)$ is a pooled hidden-state from layer $l$, such as the final prompt token representation or a mean pooled representation over selected prompt positions. The same pooling rule is used for all methods and controls. To measure hidden-state response, we add small Gaussian perturbations~\cite{goodfellow2014explaining} to the input embeddings. For perturbation index $k$, define:
\begin{equation}
\widetilde E_k(x)=E(x)+\epsilon \eta_k,
\quad
\eta_k\sim\mathcal{N}(0,I),
\end{equation}
where $\epsilon$ controls perturbation scale. Let $r_l(x;\widetilde E_k)$ denote the same pooled layer representation when the model is run with the perturbed embedding sequence $\widetilde E_k(x)$. The layerwise susceptibility is defined as:
\begin{equation}
\chi_l(x) =
\frac{1}{K}\sum_{k=1}^{K}
\left(
\frac{\|r_l(x;\widetilde E_k)-r_l(x)\|_2}
{\epsilon(\|r_l(x)\|_2+\tau)}
\right)^2 ,
\end{equation}
where $K$ is the number of perturbations and $\tau$ is a small numerical constant. We aggregate the layerwise responses over a selected layer set $\mathcal{L}_\chi$ to obtain a sample-level susceptibility:
\begin{equation}
\chi(x)=
\frac{1}{|\mathcal{L}_\chi|}
\sum_{l\in\mathcal{L}_\chi}\chi_l(x).
\end{equation}
This quantity measures response intensity. It should be interpreted as a response-intensity measure rather than a difficulty score or a direct predictor of correctness.

\subsection{Length Controlled Susceptibility}
Raw susceptibility can be affected by prompt length. To separate length effects from reasoning state effects, we fit a length response model on the calibration split within each domain:
\begin{equation}
\log(\chi(x)+\tau)
= a + b\log T(x) + \xi_x,
\end{equation}
where $T(x)$ is token length, $a$ and $b$ are fitted coefficients, and $\xi_x$ is the residual term. The length controlled susceptibility for each example is defined as:
\begin{equation}
\chi_{\mathrm{LC}}(x)
=
\log(\chi(x)+\tau)
- a - b\log T(x).
\end{equation}
A positive $\chi_{\mathrm{LC}}$ means that $x$ elicits stronger response than expected for inputs of similar length. A negative value indicates residual under activation after accounting for length.

\subsection{Cross-Layer Coordination}

Susceptibility measures response strength, but strong response alone does not guarantee coherent reasoning. We therefore measure whether hidden-state changes are directionally coordinated across layers. For a coordination layer set $\mathcal{L}_\Phi=\{l_1,\ldots,l_m\},$ define the normalized layer update:
\begin{equation}
u_l(x)=
\frac{r_l(x)-r_{l-1}(x)}
{\|r_l(x)-r_{l-1}(x)\|_2+\tau}.
\end{equation}
Cross-layer coordination~\cite{geva2021transformer} is the average pairwise cosine similarity among these updates:
\begin{equation}
\Phi(x)=
\frac{2}{m(m-1)}
\sum_{1\le i<j\le m}
u_{l_i}(x)^\top u_{l_j}(x).
\end{equation}
High $\Phi$ indicates that multiple layers move in consistent directions, which is expected when the model enters a more stable and internally coherent reasoning state.

\subsection{SPK Index and State Profiling}

The final profiling score combines length controlled response with coordination. Let $z(\cdot)$ denote standardization within the calibration split of the same domain. The length controlled SPK score is defined as:
\begin{equation}
\mathrm{SPK}_{\mathrm{LC}}(x)
=
z(\chi_{\mathrm{LC}}(x))
+ \lambda z(\Phi(x)),
\end{equation}
where $\lambda$ controls the contribution of cross-layer coordination. This score is used to construct two sets. Reasoning-active anchors $\mathcal{A}$ are examples that are solved correctly, lie near or below the behavioral boundary, and have high $\chi_{\mathrm{LC}}$ and high $\mathrm{SPK}_{\mathrm{LC}}$. Under-activated target candidates $\mathcal{U}$ are hard examples near or above the behavioral boundary with low $ \chi_{\mathrm{LC}}$. Baseline correctness is used to construct reasoning-active anchors and diagnostic plots, but not to select test-time target examples. Quantile thresholds are chosen on the calibration split and kept fixed for evaluation.

\subsection{Critical Reasoning Direction}

For each anchor $a\in\mathcal{A}$, we construct two inputs. The base input $x_a^{0}$ contains only the target problem. The active input $x_a^{+}$ contains SPK selected demonstrations followed by the same target problem. Let $r_l(x_a^{0})$ and $r_l(x_a^{+})$ be their layer representations. The unnormalized direction at layer $l$ is:
\begin{equation}
\bar v_l =
\frac{1}{|\mathcal{A}|}
\sum_{a\in\mathcal{A}}
\left(
r_l(x_a^{+})-r_l(x_a^{0})
\right).
\end{equation}
The injected direction is normalized as:
\begin{equation}
v_l =
\frac{\bar v_l}
{\|\bar v_l\|_2+\tau}.
\end{equation}
This direction represents the average latent shift from a base state toward a reasoning-active state. A stricter control replaces $x_a^{0}$ with a length matched random demonstration input, which tests whether the direction is caused by reasoning-active content rather than added context length.



\begin{table}[t]
\centering
\small
\renewcommand{\arraystretch}{1}
\setlength{\tabcolsep}{7pt}
\resizebox{\columnwidth}{!}{%
\begin{tabular}{llccc ccc cccc}
\toprule
\multirow{2}{*}{\textbf{Model}} & \multirow{2}{*}{\textbf{Method}}
& \multicolumn{3}{c}{\textbf{GSM8K}}
& \multicolumn{3}{c}{\textbf{MATH-500}}
& \multicolumn{4}{c}{\textbf{FRONTIER-4.5K}} \\
\cmidrule(lr){3-5} \cmidrule(lr){6-8} \cmidrule(lr){9-12}
& & Alpha & Accuracy & Tokens & Alpha & Accuracy & Tokens & Alpha & Sym. & Log. & Alg. \\
\midrule
\multirow{4}{*}{Qwen3-0.6B}
& \basecell{Original} & \basecell{0} & \basecell{55.8\%} & \basecell{184} & \basecell{0} & \basecell{52.0\%} & \basecell{656} & \basecell{0} & \basecell{55.5\%} & \basecell{48.2\%} & \basecell{12.5\%} \\
& Steering & 0.25 & \textbf{56.3}\% & 184 & 0.25 & 52.6\% & 662 & 0.25 & 56.2\% & 48.6\% & 14.0\% \\
& Steering & 0.5  & 56.0\% & 187 & 0.5  & 52.8\% & 645 & 0.5 & \textbf{59.2}\% & 49.1\% & 16.8\% \\
& Steering & 1.0  & 55.6\% & 186 & 1.0  & \textbf{53.4}\% & 640 & 1.0 & 58.5\% & \textbf{49.8}\% & \textbf{17.2}\% \\
\midrule
\multirow{4}{*}{Qwen3-4B}
& \basecell{Original} & \basecell{0} & \basecell{89.7\%} & \basecell{176} & \basecell{0} & \basecell{82.0\%} & \basecell{817} & \basecell{0} & \basecell{88.1\%} & \basecell{75.8\%} & \basecell{19.8\%} \\
& Steering & 0.25 & 89.9\% & 178 & 0.25 & 82.6\% & 812 & 0.25 & 88.4\% & 76.1\% & 22.0\% \\
& Steering & 0.5  & 90.2\% & 178 & 0.5  & 83.4\% & 842 & 0.5 & \textbf{92.3}\% & 76.5\% & 24.4\% \\
& Steering & 1.0  & \textbf{90.5}\% & 177 & 1.0  & \textbf{84.0}\% & 843 & 1.0 & 91.6\% & \textbf{76.8}\% & \textbf{25.7}\% \\
\midrule
\multirow{4}{*}{Qwen3-8B}
& \basecell{Original} & \basecell{0} & \basecell{92.1\%} & \basecell{213} & \basecell{0} & \basecell{82.4\%} & \basecell{896} & \basecell{0} & \basecell{88.8\%} & \basecell{76.4\%} & \basecell{20.1\%} \\
& Steering & 0.25 & 92.4\% & 215 & 0.25 & 83.6\% & 885 & 0.25 & 89.2\% & 76.9\% & 23.2\% \\
& Steering & 0.5  & 92.6\% & 213 & 0.5  & \textbf{84.2}\% & 873 & 0.5 & \textbf{93.1}\% & 77.3\% & 25.6\% \\
& Steering & 1.0  & \textbf{93.2}\% & 210 & 1.0  & 84.0\% & 902 & 1.0 & 92.3\% & \textbf{77.6}\% & \textbf{26.4}\% \\
\bottomrule
\end{tabular}}
\caption{Main results of \textsc{SPARK}-Steering on GSM8K, MATH-500, and FRONTIER-4.5K with Qwen3 series models. Shaded rows denote the original baseline without steering. For GSM8K and MATH-500, we report the steering strength $\alpha$, exact-match accuracy, and average generated tokens. For FRONTIER-4.5K, we report accuracy on symbolic (Sym.), logical (Log.), and algorithmic (Alg.) reasoning subsets. Bold numbers mark the best accuracy within each model and benchmark.}
\label{tab:main_qwen}
\vspace{-1mm}
\end{table}

\section{Experiments}
\label{sec:experiments}

We evaluate \textsc{SPARK}-Steering from three perspectives: standard benchmark performance, sensitivity to steering strength and model family, and agreement between the observed gains and the proposed latent under-activation mechanism.

\subsection{Experimental Setup}
\label{subsec:exp_setup}

\paragraph{Benchmarks and models.}
We use GSM8K and MATH-500 as reasoning benchmarks~\cite{cobbe2021training,hendrycks2021measuring,lightman2024let}. GSM8K mainly contains arithmetic problems, while MATH-500 contains more challenging problems. We also evaluate on the held-out split of our reasoning suite \textsc{FRONTIER}-4.5K. The main experiments use Qwen3 series models~\cite{yang2025qwen3}. Experiments further test Llama-3.1-8B~\cite{grattafiori2024llama} and DeepSeek-R1-Distill-Qwen-7B~\cite{guo2025deepseek}, abbreviated as DeepSeek-R1 in the following for readability.

\paragraph{Steering protocol and metrics.}
\textsc{SPARK} is train-free in the sense that no model parameters are updated and no task-specific predictor is trained. It is nevertheless data-calibrated: before evaluation, we run the frozen model on training/calibration questions to estimate length-response baselines, susceptibility distributions, reasoning-active anchor pools, and benchmark-matched steering layers. Calibration labels are used to determine correctness for anchor selection and threshold calibration, while the reported accuracies are computed on held-out evaluation questions. The \emph{Original} setting runs the frozen model without intervention. \emph{Steering} injects a \textsc{SPARK} direction extracted from reasoning-active anchors selected by length-controlled susceptibility and cross-layer coordination. We report a sweep over steering strength $\alpha$ rather than optimizing model parameters, and measure exact-match accuracy together with average generated tokens.

\subsection{Main Results}
\label{subsec:main_results}
Table~\ref{tab:main_qwen} reports the low-alpha sweep on GSM8K, MATH-500, and FRONTIER-4.5K. Across Qwen3 model sizes, moderate steering improves accuracy without noticeably increasing generation length. On GSM8K, where larger models are close to saturation, the gains are modest but stable. Qwen3-0.6B improves from 55.8\% to 56.3\%, Qwen3-4B from 89.7\% to 90.5\%, and Qwen3-8B from 92.1\% to 93.2\%. The gains are larger on MATH-500, where examples require more sustained multi-step reasoning. Qwen3-0.6B improves from 52.0\% to 53.4\%, Qwen3-4B from 82.0\% to 84.0\%, and Qwen3-8B from 82.4\% to 84.2\%. FRONTIER-4.5K shows the same trend in a controlled programmatic setting. Steering consistently improves symbolic, logical, and algorithmic reasoning across model sizes, with the largest gains on the algorithmic subset. Qwen3-0.6B improves from 12.5\% to 17.2\%, Qwen3-4B from 19.8\% to 25.7\%, and Qwen3-8B from 20.1\% to 26.4\%. Together, these results suggest that \textsc{SPARK}-Steering is most useful when the model has partial reasoning ability but the original forward pass does not enter a sufficiently active latent state. The average token counts remain in the same range, indicating that the improvement is not simply a consequence of longer generations.

\subsection{Analysis}
\label{subsec:analysis}

\paragraph{Alpha sensitivity.}
\label{subsec:alpha}
Table~\ref{tab:math_alpha} extends the MATH-500 sweep to larger steering strengths. The response is clearly non-monotonic, which is important for interpreting \textsc{SPARK} as a targeted latent intervention rather than arbitrary activation amplification. Qwen3-0.6B reaches its best score at $\alpha=1.0$ or $1.5$, improving from 52.0\% to 53.4\%. Qwen3-4B improves from 82.0\% to 84.6\%, with the best results at $\alpha=2.5$ and $3.0$. Qwen3-8B obtains the largest gain, improving from 82.4\% to 85.6\% at $\alpha=2.0$ and $\alpha=3.0$. This dose-response pattern suggests that moderate steering can move the hidden trajectory toward a more useful reasoning state, while overly strong steering can interfere with the model's native computation. The preferred steering strength also changes with model scale: larger Qwen3 models tolerate stronger intervention and obtain larger MATH-500 gains.

\begin{table}[t]
\centering
\small
\renewcommand{\arraystretch}{1}
\setlength{\tabcolsep}{7pt}
\begin{tabular}{llccc}
\toprule
\multirow{2}{*}{\textbf{Model}} & \multirow{2}{*}{\textbf{Method}}
& \multicolumn{3}{c}{\textbf{MATH-500}} \\
\cmidrule(lr){3-5}
& & Alpha & Accuracy & Tokens \\
\midrule
\multirow{4}{*}{Llama-3.1-8B}
& Original & 0    & 45.6\% & 757 \\
& Steering & 0.25 & 46.8\% & 760 \\
& Steering & 0.5  & \textbf{48.4}\% & 847 \\
& Steering & 1.0  & 47.6\% & 870 \\
\midrule
\multirow{4}{*}{DeepSeek-R1}
& Original & 0    & 85.4\% & 4138 \\
& Steering & 0.25 & \textbf{86.4}\% & 4257 \\
& Steering & 0.5  & 86.2\% & 4373 \\
& Steering & 1.0  & 32.0\% & 7519 \\
\bottomrule
\end{tabular}
\caption{Evaluation on MATH-500. The table reports original and steered performance (\%) for Llama-3.1-8B and DeepSeek-R1 under different steering strengths, including exact-match accuracy and average generated tokens.}
\label{tab:cross_family}
\vspace{-1mm}
\end{table}

\begin{table}[t]
\centering
\small
\renewcommand{\arraystretch}{1}
\setlength{\tabcolsep}{7pt}
\begin{tabular}{llccc}
\toprule
\multirow{2}{*}{\textbf{Model}} & \multirow{2}{*}{\textbf{Method}}
& \multicolumn{3}{c}{\textbf{MATH-500}} \\
\cmidrule(lr){3-5}
& & Alpha & Accuracy & Tokens \\
\midrule
\multirow{5}{*}{Qwen3-0.6B}
& Steering & 1.5 & \textbf{53.4}\% & 668 \\
& Steering & 2.0 & 52.8\% & 651 \\
& Steering & 2.5 & 52.6\% & 644 \\
& Steering & 3.0 & 52.2\% & 631 \\
& Steering & 4.0 & 52.4\% & 640 \\
\midrule
\multirow{5}{*}{Qwen3-4B}
& Steering & 1.5 & 83.8\% & 895 \\
& Steering & 2.0 & 84.0\% & 830 \\
& Steering & 2.5 & \textbf{84.6}\% & 806 \\
& Steering & 3.0 & \textbf{84.6}\% & 820 \\
& Steering & 4.0 & 84.2\% & 834 \\
\midrule
\multirow{5}{*}{Qwen3-8B}
& Steering & 1.5 & 84.4\% & 857 \\
& Steering & 2.0 & 85.6\% & 872 \\
& Steering & 2.5 & 85.4\% & 901 \\
& Steering & 3.0 & \textbf{85.6}\% & 854 \\
& Steering & 4.0 & 85.0\% & 893 \\
\bottomrule
\end{tabular}
\caption{Large-alpha sweep on MATH-500 for Qwen3 models. Each row reports the steering strength $\alpha$, exact-match accuracy (\%), and average generated tokens for a single model.}
\label{tab:math_alpha}
\end{table}

\begin{figure*}[t]
  \centering
  \includegraphics[width=1.0\textwidth]{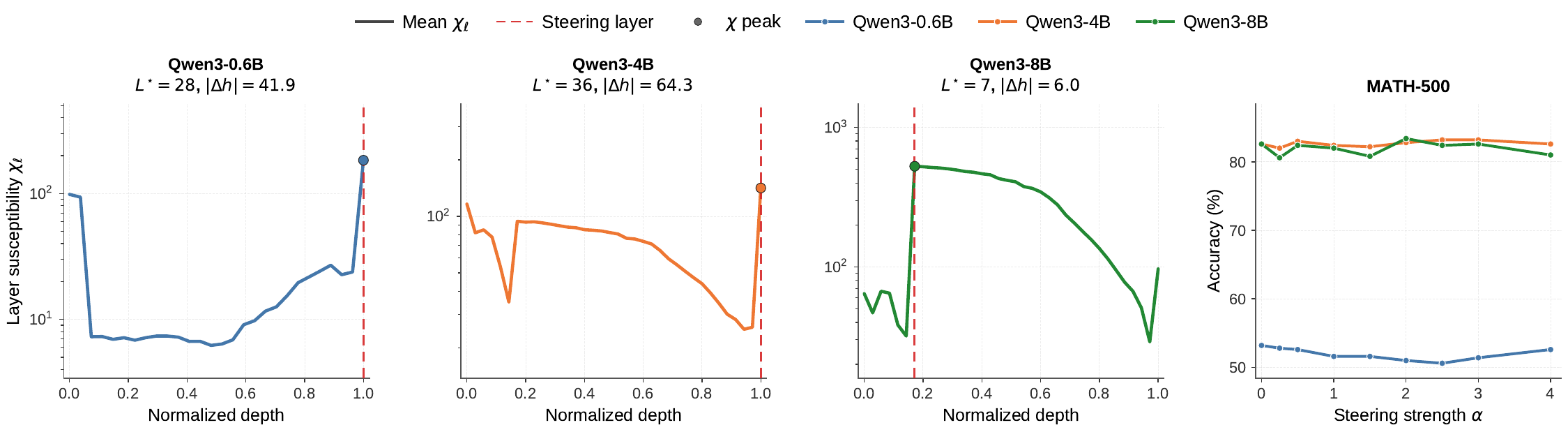}
  \caption{\textbf{Layer susceptibility and single-layer steering dynamics.} The first three plots report the layerwise susceptibility spectrum $\chi_l$ on the MATH-500 benchmark inputs. The x-axis is normalized layer depth, the y-axis is mean $\chi_l$ on a log scale, and shaded regions denote SEM. Red dashed lines mark the selected steering layer $L^\star$, while circles mark susceptibility peaks. The last plot reports held-out MATH-500 accuracy as steering strength $\alpha$ varies when injecting $h \leftarrow h+\alpha v$ at $L^\star$. The selected layers align with dominant susceptibility peaks, and the downstream accuracy changes remain bounded.}
  \label{fig:case}
\end{figure*}

\paragraph{Cross-Family Generalization}
\label{subsec:cross_family}

Table~\ref{tab:cross_family} tests whether the steering effect transfers beyond Qwen3. On Llama-3.1-8B, moderate steering improves MATH-500 accuracy from 45.6\% to 48.4\% at $\alpha=0.5$, with only a moderate increase in generated tokens, which indicates that the learned direction is not tied to a single Qwen3 model size or architecture. DeepSeek-R1 exhibits a different sensitivity profile. Small steering strengths still improve accuracy, from 85.4\% to 86.4\% at $\alpha=0.25$ and 86.2\% at $\alpha=0.5$. However, increasing $\alpha$ to 1.0 sharply reduces accuracy to 32.0\% while increasing average generation length from 4138 to 7519 tokens. We attribute this collapse to over-steering: DeepSeek-R1 already follows long native reasoning trajectories, and a large activation shift can disrupt these trajectories, making generation overly verbose and unstable.
\paragraph{Layer Susceptibility and Single-Layer Steering.}
We next examine whether the layers selected by \textsc{SPARK} correspond to meaningful susceptibility structure. Figure~\ref{fig:case} shows the layerwise susceptibility spectrum $\chi_l$ and the resulting MATH-500 accuracy under single-layer steering at the selected layer $L^\star$. Across Qwen3 model sizes, the selected steering layer consistently aligns with a dominant susceptibility peak. For Qwen3-0.6B and Qwen3-4B, $L^\star$ lies near the final layer, matching the sharp late-layer rise of $\chi_l$. For Qwen3-8B, $L^\star$ shifts to an earlier layer, suggesting that the most sensitive reasoning-related layer can move with scale. The downstream accuracy curves remain smooth and bounded as $\alpha$ varies, supporting the view that \textsc{SPARK} selects a controlled intervention site rather than injecting an arbitrary perturbation.

\begin{figure*}[t]
  \centering
  \includegraphics[width=0.85\textwidth]{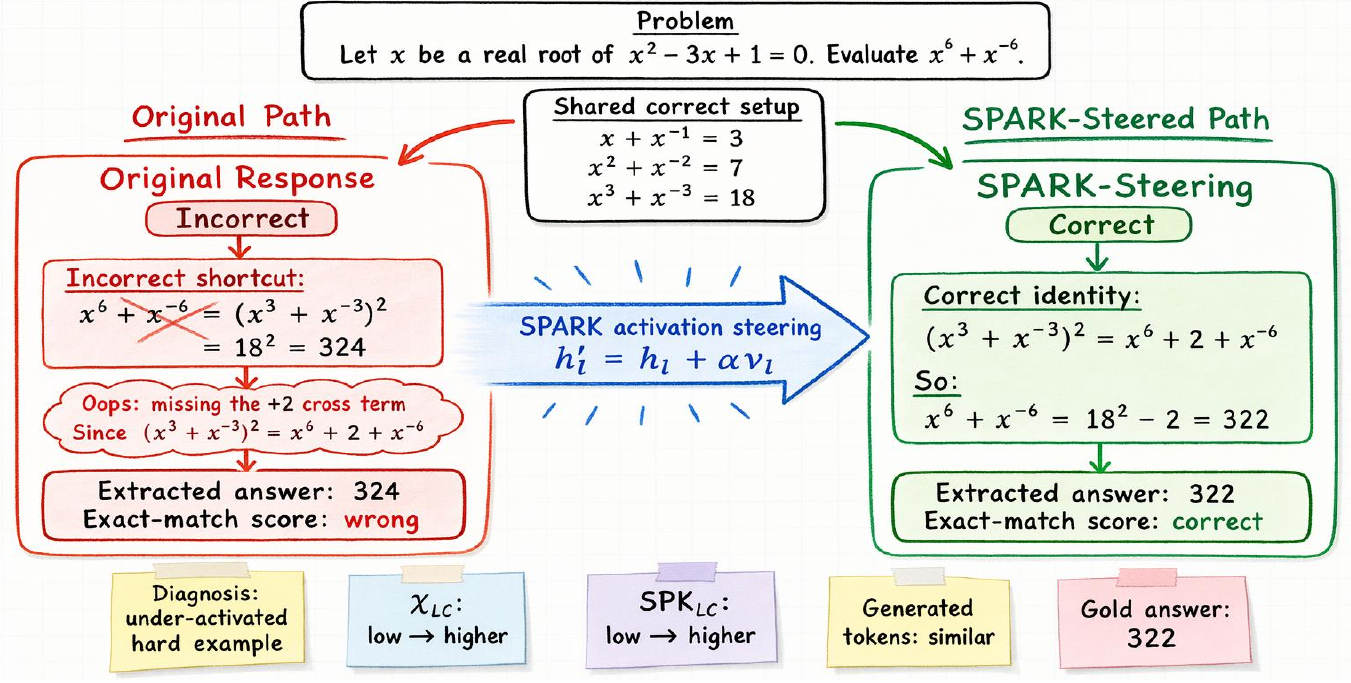}
  \caption{\textbf{Qualitative case study.} The base model reaches the correct setup but makes a local algebraic error when converting $(x^3+x^{-3})^2$ into $x^6+x^{-6}$. \textsc{SPARK}-Steering preserves the setup and repairs this local transformation, changing the
  extracted answer from $324$ to the gold answer $322$. The example is scored by exact match after final-answer extraction.}
  \label{fig:case_study}
  \vspace{-1mm}
\end{figure*}

\subsection{Case Study}
\label{subsec:case_study}

Figure~\ref{fig:case_study} shows a representative algebra example where the base model reaches the correct setup but fails at a local transformation. Given a real root of $x^2-3x+1=0$, the task is to compute $x^6+x^{-6}$. Both the original and steered responses correctly derive $x+x^{-1}=3$ and $x^3+x^{-3}=18$. The original response then incorrectly sets $x^6+x^{-6}=(x^3+x^{-3})^2=324$, omitting the cross term in $(x^3+x^{-3})^2=x^6+2+x^{-6}$. With \textsc{SPARK}-Steering at $\alpha=3.0$, the model preserves the setup but applies the corrected identity, producing $18^2-2=322$. The extracted final answer changes from $324$ to the gold answer $322$, and the example is scored after final-answer extraction.

\subsection{Ablation Studies}
\label{subsec:ablations}

\begin{table}[t]
\centering
\small
\renewcommand{\arraystretch}{1}
\setlength{\tabcolsep}{6pt}
\begin{tabular}{llcc}
\toprule
\textbf{Ablation} & \textbf{Setting} & \textbf{Accuracy} & $\boldsymbol{\Delta}$ \\
\midrule
Baseline & Original & 82.4\% & 0.0 \\
\midrule
\multirow{5}{*}{Direction}
& Random direction & 82.6\% & +0.2 \\
& Easy correct anchors & 83.2\% & +0.8 \\
& High raw-$\chi$ anchors & 83.8\% & +1.4 \\
& High $\chi_{\mathrm{LC}}$ anchors & 84.8\% & +2.4 \\
& High $\mathrm{SPK}_{\mathrm{LC}}$ anchors & \textbf{85.6}\% & \textbf{+3.2} \\
\midrule
\multirow{5}{*}{Layer}
& Early layer & 84.2\% & +1.8 \\
& Middle layer & 84.6\% & +2.2 \\
& Late layer & 85.0\% & +2.6 \\
& $\chi_l$ peak layer & 85.4\% & +3.0 \\
& SPARK-selected layer & \textbf{85.6}\% & \textbf{+3.2} \\
\bottomrule
\end{tabular}
\caption{Ablation study on MATH-500 with Qwen3-8B. $\Delta$ denotes the accuracy change over the original model.
Direction ablations test how the steering vector is constructed, while layer ablations test where the vector is
injected.}
\label{tab:ablation}
\vspace{-1mm}
\end{table}

Table~\ref{tab:ablation} compares different direction-construction and layer-selection choices. Random directions give only a small gain, indicating that arbitrary activation noise does not explain the improvement. Easy correct anchors and high raw-$\chi$ anchors improve over the baseline, but remain weaker than length-controlled directions. This supports the central role of length control: raw susceptibility captures useful response information, but it is still confounded by prompt length.
The best direction is obtained from high $\mathrm{SPK}_{\mathrm{LC}}$ anchors, which combine length-controlled response with cross-layer coordination. Layer ablations show the same pattern. Fixed early, middle, or late layers provide partial gains, but susceptibility-guided layer selection performs best.

\section{Conclusion}
This work studies reasoning failures in LLMs through the lens of latent state activation. Final-answer accuracy reveals where a model succeeds or fails, but it does not distinguish missing capability from a failure to activate an available reasoning state. We show that hidden-state susceptibility offers a useful internal response signal for this distinction, while also revealing an important caveat. Raw susceptibility is strongly affected by prompt length, especially in algorithmic reasoning. Length-controlled susceptibility addresses this confound and provides a cleaner view of residual reasoning activation in hard examples.
Building on this diagnosis, \textsc{SPARK} connects latent profiling with test-time intervention. It identifies reasoning-active anchors, detects under-activated hard targets, and uses the diagnosed active direction to steer hidden states while keeping model parameters frozen. Experiments on standard benchmarks show that \textsc{SPARK} improves reasoning performance, with stronger gains on harder examples and limited disruption to easier ones. These findings suggest that reasoning evaluation should not stop at measuring whether a model answers correctly. A more informative evaluation also asks whether the internal computation needed for reasoning is activated, how this activation is shaped by input scale, and whether it can be guided during inference.


\clearpage
\newpage
\bibliographystyle{plainnat}
\setcitestyle{numbers}
\bibliography{ref}



\end{document}